\documentclass[sigconf]{acmart}

\usepackage{multirow}
\usepackage{colortbl}
\usepackage{makecell}
\usepackage{pifont}
\usepackage{xcolor}
\usepackage[table]{xcolor}
\definecolor{myyellow1}{RGB}{254,241,206}
\definecolor{mygreen1}{RGB}{220,241,217}
\definecolor{myblue1}{RGB}{217,231,247}
\AtBeginDocument{%
  }
\copyrightyear{2026}
\acmYear{2026}
\setcopyright{cc}
\setcctype{by}
\acmConference[MM '26]{Proceedings of the 34th ACM International Conference on Multimedia}{November 10--14, 2026}{Rio de Janeiro, Brazil}
\acmBooktitle{Proceedings of the 34th ACM International Conference on Multimedia (MM '26), November 10--14, 2026, Rio de Janeiro, Brazil}
\acmDOI{10.1145/3767308.3835380}
\acmISBN{979-8-4007-2213-4/2026/11}
\begin{document}

\title{Erase but Preserve: Controllable Removal of Copyrighted Animation Characters via Optimized Semantic Anchors}

\author{Qiao Li}
\email{liqiao@iie.ac.cn}
\orcid{0009-0004-1915-2570}
\affiliation{%
  \institution{Institute of Information Engineering, Chinese Academy of Sciences\\
    School of Cyber Security, University of Chinese Academy of Sciences}
  \city{Beijing}
  \country{China}
}

\author{Xiaomeng Fu}
\email{fuxiaomeng@iie.ac.cn}
\orcid{0000-0001-7195-0765}
\affiliation{%
  \institution{Institute of Information Engineering, Chinese Academy of Sciences\\
    School of Cyber Security, University of Chinese Academy of Sciences}
  \city{Beijing}
  \country{China}
}

\author{Wangjia Yu}
\email{yuwangjia@iie.ac.cn}
\orcid{0009-0001-5777-4344}
\affiliation{%
  \institution{Institute of Information Engineering, Chinese Academy of Sciences\\
    School of Cyber Security, University of Chinese Academy of Sciences}
  \city{Beijing}
  \country{China}
}

\author{Runze He}
\email{hrz010109@gmail.com}
\orcid{0009-0009-7917-7223}
\affiliation{%
  \institution{Institute of Information Engineering, Chinese Academy of Sciences\\
    School of Cyber Security, University of Chinese Academy of Sciences}
  \city{Beijing}
  \country{China}
}

\author{Baisen Wang}
\email{wbs2788@gmail.com}
\orcid{0000-0001-5137-9709}
\affiliation{%
  \institution{Institute of Information Engineering, Chinese Academy of Sciences\\
    School of Cyber Security, University of Chinese Academy of Sciences}
  \city{Beijing}
  \country{China}
}

\author{Jiao Dai}
\correspondingauthor
\email{daijiao@iie.ac.cn}
\orcid{0000-0003-3559-8009}
\affiliation{%
  \institution{Institute of Information Engineering, Chinese Academy of Sciences}
  \city{Beijing}
  \country{China}
}

\author{Jizhong Han}
\email{hanjizhong@iie.ac.cn}
\orcid{0000-0003-1107-3873}
\affiliation{%
  \institution{Institute of Information Engineering, Chinese Academy of Sciences}
  \city{Beijing}
  \country{China}
}

\renewcommand{\shortauthors}{Qiao Li et al.}

\begin{abstract}
  The exceptional generation capabilities of text-to-image diffusion models have raised copyright concerns, particularly the unauthorized reproduction of animation characters. Existing concept erasure methods fall short for animation character erasure: model modification methods struggle to identify suitable anchors for diverse, highly distinctive characters; prompt-based steering methods lack fine-grained control for precise intervention. These approaches often yield incomplete erasure and degraded image fidelity, hindering real-world deployment. In this paper, we propose a controllable method operating on the model’s continuous textual representation to erase target characters during generation. We optimizes an anchor embedding via structural and detailed constraints to serve as a character surrogate, then replaces target-related embeddings with the anchor via a structure-aware adaptive strategy. Experiments show that our method achieves state-of-the-art erasure effectiveness and image fidelity preservation, while supporting controllable erasure degree, multi-target removal, and model transferability. Moreover, our optimized anchors are plug-and-play with current model modification baselines to improve their erasure performance.
\end{abstract}

\ccsdesc[500]{Security and privacy~Human and societal aspects of security and privacy}
\ccsdesc[300]{Computing methodologies~Computer vision}

\keywords{Copyright protection; Concept erasure; Diffusion models}
\begin{teaserfigure}
  \centering
  \includegraphics[width=0.80\textwidth]{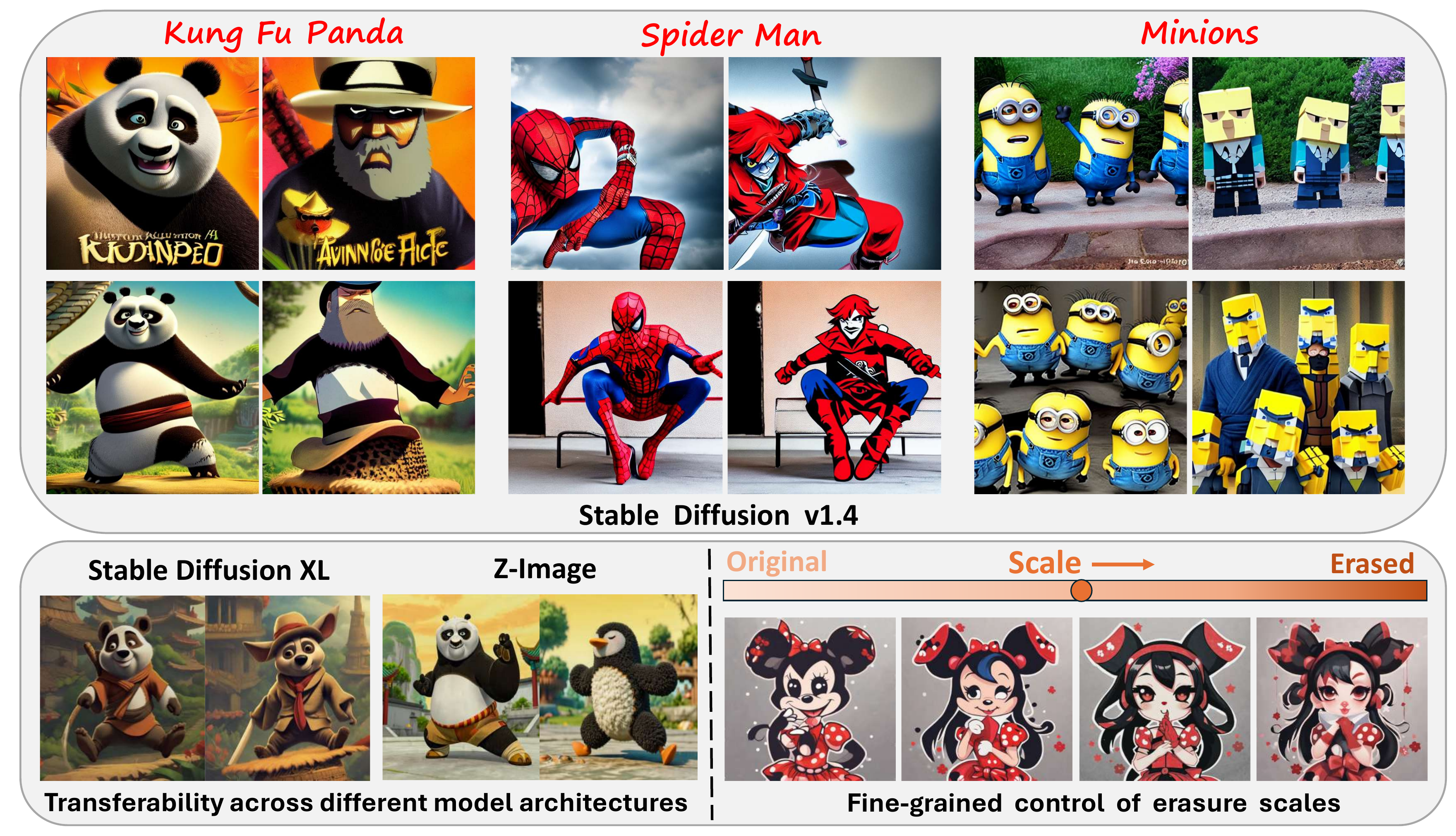}
  \caption{During image generation, our method effectively erases diverse animation concepts using optimized semantic anchors, while preserving overall image fidelity. It also supports model transferability and fine-grained control over the erasure scale.}
  \label{fig:teaser}
\end{teaserfigure}


\maketitle

\begin{figure*}[htbp]
    \centering
    \includegraphics[width=0.98\textwidth]{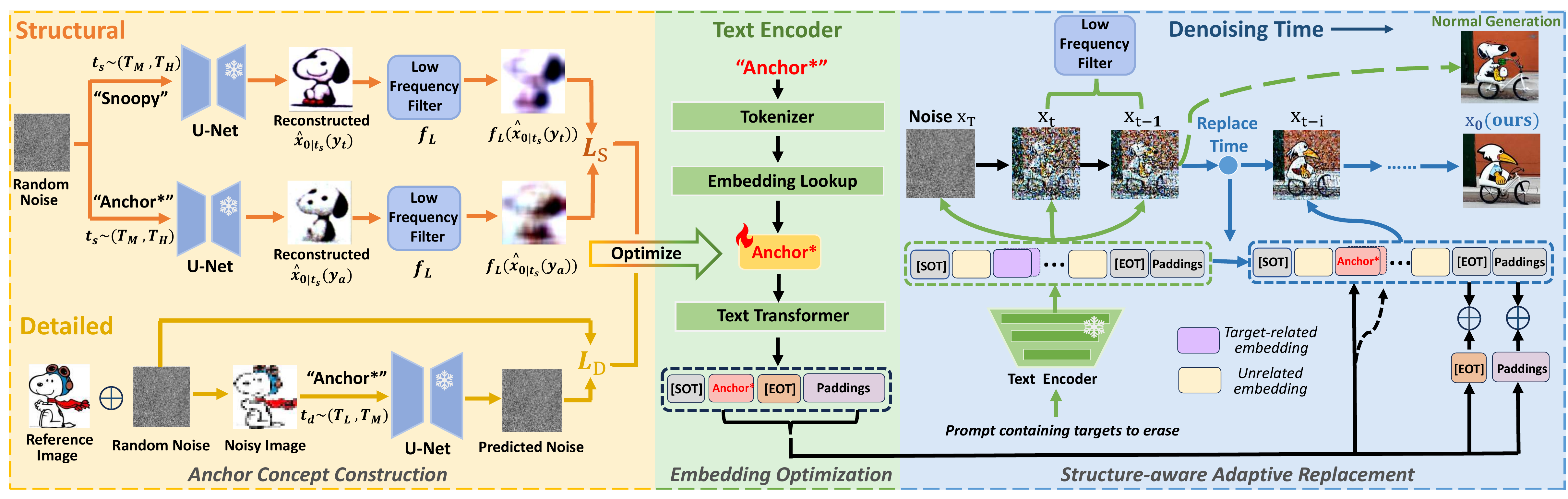} 
    \caption{The overall pipeline of our proposed method. We construct an anchor by applying structural constraints (top-left) and detailed constraints (bottom-left), and optimize an anchor embedding in the continuous textual embedding space (middle). During inference, when the input prompt contains target-related terms in the predefined subspace, our method performs structure-aware adaptive embedding replacement to erase the target using the optimized anchor concept (blue line on the right), compared with normal generation (green line on the right).}
    \label{fig:method}
\end{figure*}

\section{Introduction}
Text-to-image diffusion models~\cite{Song2019GenerativeMB,Ho2020DenoisingDP,Rombach2021HighResolutionIS} have become a core tool for visual content creation, routinely used in advertising, filmmaking, and user-generated content platforms. However, this broad adoption raises legal and ethical risks~\cite{growcoot2022midjourney,Zhang2023OnCR,Lu2024DisguisedCI,bozard2024does}, as these models can generate unsafe or infringing content that violates policies. A practical and high-impact case arising from commercial deployment is the unauthorized generation of copyrighted animation characters. Recent disputes, such as the lawsuit by Disney and Universal Studios against Midjourney regarding images resembling characters like Spider-Man and the Minions~\cite{bbc_news}, highlight that this issue is not hypothetical: it directly affects product deployment, platform governance, and creators who rely on generative models. 

To mitigate the risks of undesired generation, concept erasure techniques have emerged as one of the feasible solutions, primarily aiming to prevent models from generating unsafe concepts. Existing erasure approaches can be categorized into: \textbf{(i) model modification} methods that alter model parameters to erase or suppress undesired concepts, typically by mapping them to a neutral or benign anchor concept~\cite{Gandikota2023UnifiedCE,Lyu2023OnedimensionalAT,Kumari2023AblatingCI,Zhang2023ForgetMeNotLT,Lu2024MACEMC,Gong2024ReliableAE,Bui2024ErasingUC}, and \textbf{(ii) prompt-based steering} methods that adjust input prompts or introduce negative terms to avoid undesired concepts during inference~\cite{Schramowski2022SafeLD,Jain2024TraSCETS,DBLP:conf/iclr/YoonYPYB25,na2025trainingfree}. Although effective in certain scenarios (e.g. erasing unsafe concepts like ``nudity''), neither approach adequately focuses on and addresses the specific challenges of erasing copyrighted animation characters in real-world deployment, primarily due to the unique properties of these characters. 

First, animation characters exhibit a large variety, with each being highly distinctive. This poses challenges for model modification methods, which typically require selecting an appropriate anchor as a benign surrogate for undesired target. While existing anchor selections such as synonyms, parent/child classes, or general concepts (e.g., null text or ``ground'') work well for generic categories (e.g., mapping ``grumpy cat'' to ``cat'', ``nudity'' to ``clothed''), they are often ill-suited for specific characters. Identifying appropriate semantic synonyms or taxonomic relations for these unique characters is often laborious or even infeasible. Besides, prior studies~\cite{Zhang2025BeyondFA,Lyu2023OnedimensionalAT} indicate that resorting to general or semantically distant anchors can significantly compromise erasure performance, including incomplete erasure and context contamination.

Second, erasing a copyrighted animation character often requires more nuanced intervention than coarsely blocking an entire class of not-safe-for-work (NSFW) content. (i) Unlike inherently harmful NSFW content, the assessment of animation infringement varies across laws and platform policies, and may in some cases permit moderate visual similarity that does not constitute copyright infringement. (ii) Animation imagery often carries commercial and entertainment value on user‑generated content platforms. Instead of indiscriminately blocking all potentially infringing prompts, platforms typically aim to preserve the user's original creative intent (e.g., composition, style, background, unrelated elements) while only excising the copyrighted characters. These two concerns necessitate fine-grained control over both the erased character subject and the preserved contextual elements during inference. However, existing prompt-based steering methods typically rely on discrete textual descriptions that provide only coarse control over generation, thereby lacking precise regulation of both the character erasure degree and the surrounding context retention.

To address these challenges, we propose a method that controllably erases animation characters during generation by operating on the model’s continuous textual representation. Our key idea is to replace the target character with a learned \emph{anchor} concept that explicitly erases its primary visual features while preserving unrelated contextual elements from the prompt. Specifically, we first \textbf{optimize an anchor embedding} by extracting structure outlines and detailed features from the target's visual semantics. This learned anchor is then used to replace the target to guide generation toward a non-infringing surrogate. As the anchor is represented in a continuous embedding space, our method enables fine-grained control over the character erasure degree via adjustment of the replacement intensity. To avoid unintended alterations to unrelated context, we perform \textbf{targeted embedding replacement}: leveraging the disentanglement property of textual embeddings, we replace only the embeddings related to the target character while retaining unrelated elements. This replacement is applied \textbf{adaptively across denoising timesteps}, which further improves both reliable target removal and overall scene coherence.

Due to the lack of standard benchmark for animation character erasure, we build a dataset of 80 animation concepts that can be reliably generated by diffusion models. Experiments show that our approach achieves superior performance in both target character removal and overall fidelity preservation compared with baselines. Our method also supports fine-grained control over the erasure degree, simultaneous removal of multiple targets, and transferability across different diffusion models (including models with dual text encoders and recent DiT-based models). Moreover, our optimized anchors can also be directly integrated into current model modification methods as a benign surrogate. Experiments demonstrate that, compared to adopting existing general anchors, leveraging our learned anchors yields higher erasure accuracy and improved image fidelity preservation in animation character erasure.

Our contributions are summarized as:
\begin{itemize}
    \item  We propose a novel method to erase one or more copyrighted animation characters directly during the generation process of text-to-image diffusion models.
    
    \item  Our continuous anchor optimization approach ingeniously leverages the visual features of target characters, offering a principled way to identify a controllable anchor for distinctive animation characters.

    \item  Our structure-aware adaptive replacement strategy jointly achieves precise target character removal and high fidelity preservation, ensuring the coherence and usability of the resulting animation imagery.

    \item Experiments show that our method achieves state-of-the-arts in animation character erasure, while enabling controllable erasure degree, simultaneous removal of multi-targets, and model transferability. Moreover, our optimized anchors are plug-and-play with model modification methods to improve their erasure performance.

\end{itemize}

\section{Related Work}
\label{sec:relatedwork}

\subsection{Text-to-image Diffusion Models}
Text-to-image diffusion models have garnered substantial attention due to their capacity for high-fidelity image synthesis~\cite{Dhariwal2021DiffusionMB,Ramesh2022HierarchicalTI,Saharia2022PhotorealisticTD,Balaji2022eDiffITD}. They incorporate image encoder-decoder frameworks to efficiently conduct the diffusion and denoising process within a latent space.

During the training process, random Gaussian noise $\epsilon$ is introduced to the image $x_0$:
\begin{equation}
\label{eq:1}
    x_t=\sqrt{\bar{\alpha}_t}x_0+\sqrt{1-\bar{\alpha}_t}\epsilon
\end{equation}
The training goal of diffusion models is to learn to predict the introduced noise from $x_t$ at time step t:
\begin{equation}
\label{eq:training target}
    \mathcal{L}:=\mathbb{E}_{\epsilon\sim \mathcal{N}(0,1), t\sim U(0,T)}[||\epsilon-\epsilon_{\theta}\big(x_t,t,c_{\theta}(y)\big)||_2^2]
\end{equation}
where $\epsilon_{\theta}$ is a U-Net, $c_{\theta}$ is a text encoder, y is a textual input.

During the inference process, previous works~\cite{Kwon2022DiffusionMA,Wang2023ExploitingDP,Park2024ExplainingGD} suggest that models focus on constructing low-level structure and outlines in the early denoising stages, and subsequently shift to predicting semantic details in the later stages.

\noindent\textbf{Deterministic DDIM Scheduler.} To accelerate the denoising process, deterministic DDIM sampling~\cite{Song2020DenoisingDI} has been proposed, enabling a skip-step strategy. The skip-step denoising process for any timestep $s<k$ can be mathematically formulated as follows:
\begin{equation}
\label{eq:2}
    x_s=\sqrt{\bar{\alpha_s}}{\hat{x}}_{0|k}+\sqrt{1-\bar{\alpha_s}}\epsilon_{\theta}\big(x_k,k,c_{\theta}(y)\big)
\end{equation}
where: 
\begin{equation}
\label{eq:3}
    {\hat{x}}_{0|k}=\frac{1}{\sqrt{\bar{\alpha}_k}}\Big(x_k-\sqrt{1-\bar{\alpha_k}}\epsilon\big(x_k,k,c_{\theta}(y)\big)\Big)
\end{equation}


\begin{figure}[h!]
    \centering
    \includegraphics[width=0.98\columnwidth]{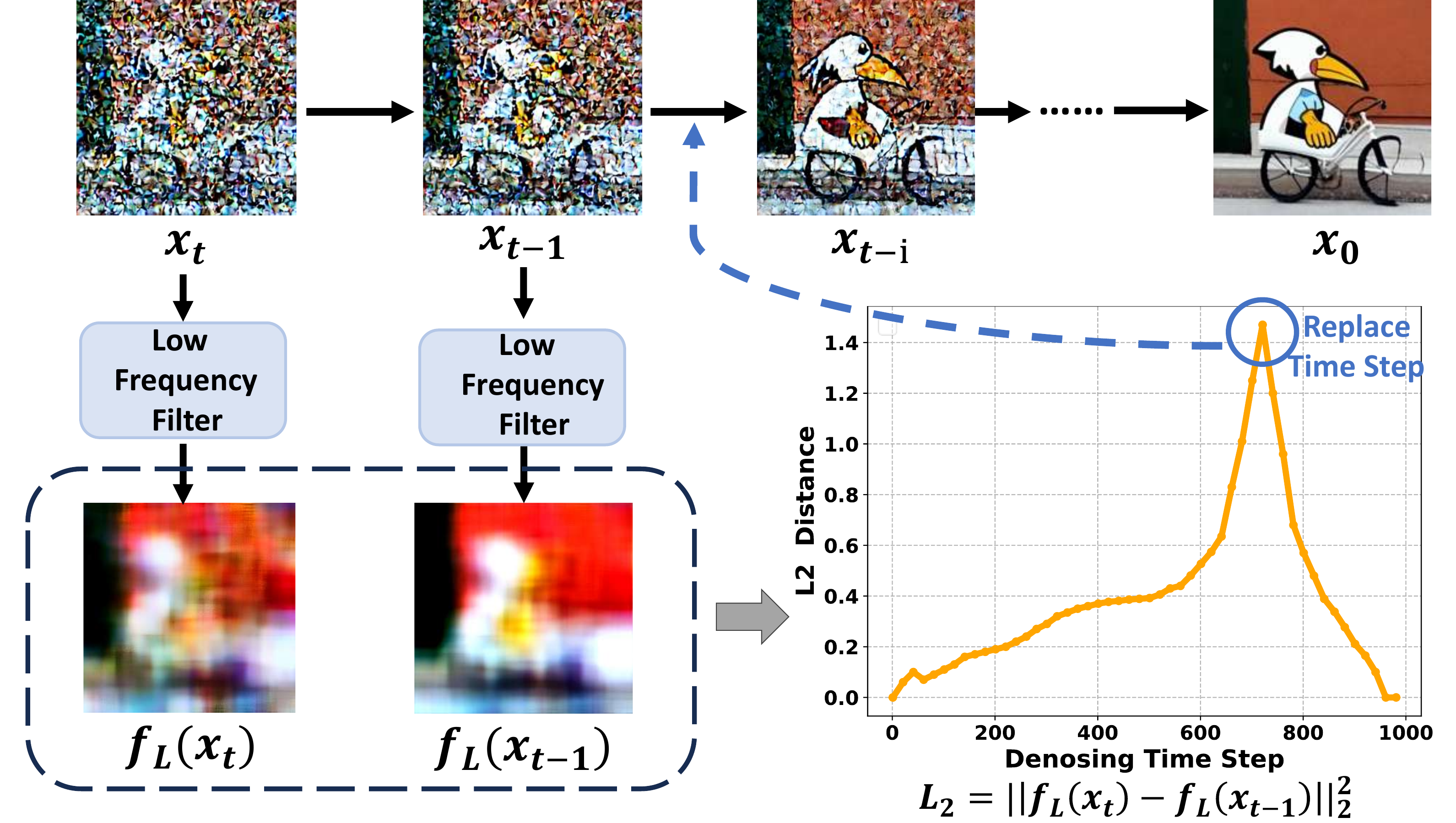} 
    \caption{Illustration of the structure-aware adaptive replacement module. We extract low-frequency structural components and analyze the changes between adjacent denoising steps to find an optimal starting point for replacement.}
    \label{fig:adaptive_replace}
\end{figure}

\subsection{Concept Erasure in Diffusion Models}
\noindent\textbf{Model Modification Methods.} Model modification methods update diffusion models' weights to either suppress the undesired target or map it onto a neutral or benign anchor concept. Most existing works~\cite{Gandikota2023UnifiedCE,Lyu2023OnedimensionalAT,Kumari2023AblatingCI,Zhang2023ForgetMeNotLT,Lu2024MACEMC,Bui2024ErasingUC} modify parameters in the cross-attention mechanism, text encoder, or the entire model to erase target concepts through iterative fine-tuning. Several works~\cite{Li2025SPEEDSP,Gong2024ReliableAE} directly derive the updated weights via closed-form solution, thus avoiding the need for fine-tuning. However, most of these methods require an appropriate anchor concept to replace the target concept. While this is relatively straightforward for generic objects (e.g. dog) or NSFW content (e.g. nudity), it can be laborious or even infeasible for various highly unique animation characters. Our method provide an effective solution for constructing a suitable anchor, which can be directly applied to existing model modification methods.

\noindent\textbf{Prompt-based Steering Methods.} Current prompt-based steering methods primarily rely on classifier-free guidance (CFG)~\cite{Ho2022ClassifierFreeDG}. Safe Latent Diffusion (SLD)~\cite{Schramowski2022SafeLD} uses multiple noise predictions to steer the unconditional prediction towards a safe prompt while avoiding the negatives. Negative Prompting 
(NP) is a technique that replaces the empty prompt in CFG with a negative one. TraSCE~\cite{Jain2024TraSCETS} modifies NP by preserving part of the unconditional predictions and introducing a loss-based guidance mechanism. SAFREE~\cite{DBLP:conf/iclr/YoonYPYB25} steers prompt tokens away from a toxic subspace. However, these methods mainly focus on global NSFW removal, which fail to provide fine-grained control, proving inadequate for animation characters that require nuanced processing.

\section{Method}
Our goal is to erase target animation characters during the generation process of a diffusion model, while preserving the overall visual fidelity of the generated images. First, we define the objective to construct an anchor concept using the target character's structural and detailed features (Section~\ref{sec:anchor}). Next, we optimize the anchor embedding in the continuous textual embedding space (Section~\ref{sec:optimization}). During inference, we selectively replace the target-related embeddings with the optimized embedding following structure-aware adaptive strategy, thereby achieving precise target erasure while preserving scene coherence (Section~\ref{sec:adaptive}). The overall pipeline of our method is illustrated in Figure~\ref{fig:method}.

\subsection{Anchor Concept Construction}
\label{sec:anchor}
We aim to remove a copyrighted animation character during generation while preserving overall image fidelity, including the coherence of the background and other contextual elements. To achieve this, we first construct an anchor concept that serves as a benign surrogate for the copyrighted target. The anchor must simultaneously fulfill two criteria: (i) its outline and general structure should be roughly similar to those of the target to ensure harmonious replacement and avoid background distortion. We define the loss function for constructing the structural outline as $\mathcal{L}_S$; (ii) its main detailed features should exhibit significant distinctiveness from those of the target to avoid copyrighted appearance. We define the loss function for differentiating details as $\mathcal{L}_D$.

Overall, we formulate the anchor construction problem as:
\begin{equation}
\label{eq:loss}
\min \left( \alpha \cdot\mathcal{L}_S + \beta \cdot \mathcal{L}_D \right)
\end{equation}
where $\alpha$ and $\beta$ are determined empirically. 

During optimization, we use the string ``Anchor*'' to represent the new anchor concept's name in the word space, as it is not defined in the text encoder's vocabulary before.

\noindent \textbf{Structural Outline Construction.} We aim to construct an anchor concept that shares a similar structural outline with the target. We leverage the model's generative prior to capture diverse structural poses and layouts. As shown in Figure~\ref{fig:method} (top-left), given a random Gaussian noise $x_T\sim \mathcal{N}(0, I)$,  we first randomly sample a large timestep $t_s\sim(T_M, T_H)$, where the diffusion model predominantly captures the global structure. We then obtain an intermediate latent $x_{t_s} (t_s<T)$ via the deterministic DDIM skip-step denoising formula (defined in Equation~\ref{eq:2}):
\begin{equation}
    x_{t_s}=\sqrt{\bar{\alpha}_{t_s}}\hat{x}_{0|T}+\sqrt{1-\bar{\alpha}_{t_s}}\epsilon_{\theta}\big(x_T,T,c_{\theta}(y_t)\big)
\end{equation}
where $\hat{x}_{0|T}$ can be derived following Equation~\ref{eq:3}, $y_t$ denotes the name of the target character (e.g. ``Snoopy''). This $x_{t_s}$ encodes the coarse structure of the target character.

Subsequently, following Equation~\ref{eq:3}, we reconstruct two original samples $\hat{x}_{0|t_s}(y)$ from $x_{t_s}$:
\begin{equation}
    \hat{x}_{0|t_s}(y)=\frac{1}{\sqrt{\bar{\alpha}_{t_s}}}\Big(x_{t_s}-\sqrt{1-\bar\alpha_{t_s}}\epsilon_{\theta}\big(x_{t_s},{t_s},{c_{\theta}(y)}\big)\Big)
\end{equation}
By applying two different prompts $y$, we can obtain two reconstructed samples: $\hat{x}_{0|t_s}(y_t)$ for the target prompt $y_t$, and $\hat{x}_{0|t_s}(y_a)$ for the anchor name $y_a$ (i.e. ``Anchor*'').

To ensure the anchor learns a similar structural outline as the target, we maximize the structural similarity between two reconstructed samples. Since structural information is typically represented by low-frequency signals, we employ a low-frequency filter $f_L$ to extract their low-frequency components $x_L(y_t)$ and $x_L(y_a)$:
\begin{equation}
    x_{L}(y_t)=f_L\big(\hat{x}_{0|t_s}(y_t)\big),  \> \> \> x_{L}(y_a)=f_L\big(\hat{x}_{0|t_s}(y_a)\big)
\end{equation}

Our goal of optimizing the anchor's overall structural outline can thus be formulated as:
\begin{equation}
    \mathcal{L}_S = \mathbb{E} \left\| x_{L}(y_t)-x_{L}(y_a) \right\|_2^2
\end{equation}

\noindent \textbf{Detailed Features Differentiation.} To ensure the erasure of infringing elements, the main detailed features of the anchor concept should differ from those of the target. We choose a clean reference image $x_0$ of the target character whose content clearly defines the infringing features to be erased. As shown in Figure~\ref{fig:method} (bottom-left), we add a Gaussian noise $\epsilon\sim \mathcal{N}(0, I)$ to $x_0$ at a random timestep $t_d \sim (T_L, T_M)$ to obtain $x_{t_d}$, as in Equation~\ref{eq:1}; this noise level primarily degrades fine details while preserving coarse structure. Based on the diffusion training objective (Equation~\ref{eq:training target}), we maximize the noise prediction error under the anchor prompt $y_a$ to prevent it from reconstructing these details: 
\begin{equation}
    \mathcal{L}_D = -\mathbb{E} \left\| \epsilon - \epsilon_{\theta}\big(x_{t_d}, t_d, c_{\theta}(y_a)\big) \right\|_2^2
\end{equation}

\begin{table*}[]
\caption{Quantitative comparison with baselines on erasing 80 characters when generating images from Stable Diffusion-v1-4. ↑ represents that a higher value indicates better performance, and vice versa. (Bold: best. Underline: second-best.)}

\renewcommand{\arraystretch}{1.0}
\resizebox{0.9\linewidth}{!}{
\begin{tabular}{cccccccc}
\toprule[1.05pt]
\multirow{2}{*}{\textbf{Method}}  & \multicolumn{2}{c}{\textbf{Erasure Effectiveness}}                            & \multicolumn{3}{c}{\textbf{Image Fidelity Preservation}}               & \multicolumn{2}{c}{\textbf{Unrelated Image}}   \\ 
 \cmidrule(lr){2-3} \cmidrule(lr){4-6} \cmidrule(lr){7-8}
         & LLaVA-1.5↓           & BLIP-3↓             & SSIM↑           & LPIPS↓         & Aesthetic↑    &          FID↓& CLIP↑                \\ \specialrule{0.8pt}{0.5ex}{0.5ex}
\multicolumn{1}{c}{SD v1.4 (Base)}      & 66.9\%& 64.7\%& 1& 0& 5.30& 33.7& 0.326\\ \hline
 SLD-medium& 42.3\%& 50.3\%& 0.314& 0.678& \underline{5.15}& \underline{34.9}&0.305\\
SLD-strong      & 20.0\%               & 18.2\%              & 0.286           & 0.707          & 5.08          & 36.1                 & 0.298                \\
SAFREE          & 12.8\%               & 11.9\%& 0.213& 0.772& 5.10& 35.3& \underline{0.307}\\
Negative Prompt & 11.1\%               & 11.5\%              & 0.384           & 0.639          & 5.10          & 35.4                 & 0.302                \\
 STG& 19.3\%& 17.0\%& \underline{0.431}& \underline{0.560}& 4.87& 38.7&0.281\\
TraSCE& \underline{9.5\%}& \underline{5.7\%}& 0.347& 0.693& 4.99& 35.6& 0.299\\ \rowcolor{blue!10}
Ours            & \textbf{6.0\%}       & \textbf{4.0\%}      & \textbf{0.467}  & \textbf{0.505} & \textbf{5.18} & \textbf{33.2}& {\textbf{0.312}}      \\ 
\bottomrule[1.1pt]
\end{tabular}}
\label{tab:baselines experiment}
\end{table*}

\subsection{Anchor Embedding Optimization}
\label{sec:optimization}
After defining the anchor concept's construction objective, we optimize it as a textual embedding in the continuous embedding space.

\noindent \textbf{Initialization.} To represent the anchor concept, we initialize a word vector $v^*\in \mathbb{R}^{1 \times D}$ ($D$ is the feature dimension) by looking up the token ``Anchor*'' in the CLIP~\cite{Radford2021LearningTV} text encoder's embedding layer. This yields a learnable starting point for anchor optimization.

\noindent \textbf{Optimization}. Following the anchor construction objective in Equation~\ref{eq:loss}, $v^*$ is optimized by minimizing:
\begin{equation}
    v^* = \arg\min_{v^*} \left(\alpha\cdot \mathcal{L}_S + \beta \cdot \mathcal{L}_D \right)
\end{equation}
During optimization, when inputting anchor prompt, we form a token sequence containing Start-of-Text (SOT), ``Anchor*'', End-of-Text (EOT), and Padding tokens. The anchor token uses current $v^*$, while other tokens use their fixed predefined vectors. After positional encoding, the sequence is passed through the text encoder's frozen Transformer layer to produce contextualized embeddings $e$. These embeddings condition the diffusion model when computing $\mathcal{L}_S$ and $\mathcal{L}_D$, and gradients are backpropagated to update $v^*$.

After optimization, we obtain the final vector $v^*$ representing the anchor concept, along with its corresponding contextualized embeddings:
\begin{equation}
\label{eq:optimized embedding}
    e^* = \{e^*_{SOT}, \> e^*_{anchor},\> e^*_{EOT},\> e^*_{Paddings}\}
\end{equation}
where $e^*_{anchor}$ is the optimized anchor embedding, $e^*_{SOT}$, $e^*_{EOT}$, and $e^*_{Paddings}$ denote SOT, EOT, and Padding embeddings, respectively.

\begin{figure*}[h!]
    \centering
    \includegraphics[width=0.8\linewidth]{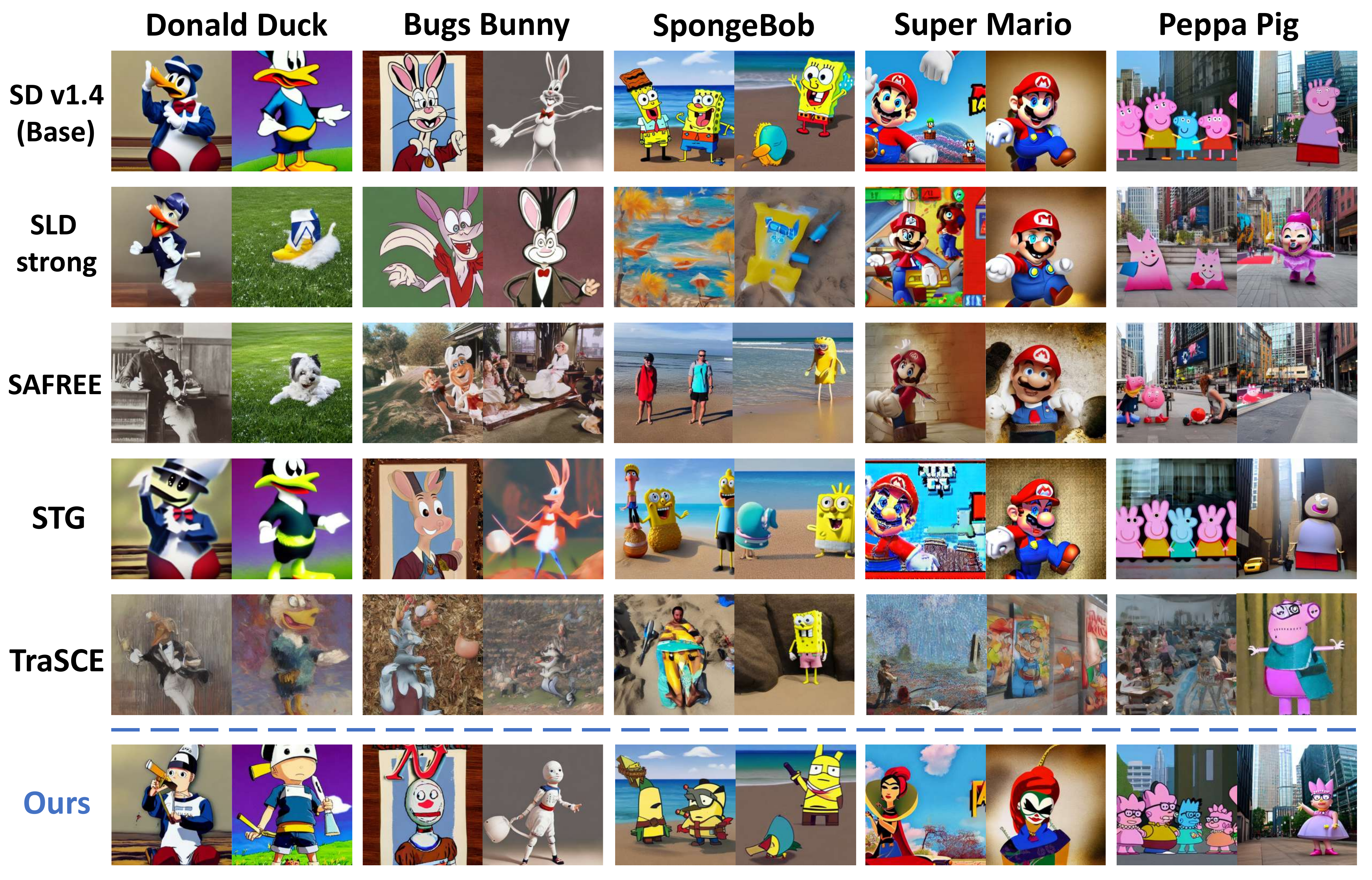} 
    \caption{Qualitative comparison between our method and baseline methods. Our method completely erases target animation concepts with optimized anchor concepts, while better preserving overall image fidelity and background coherence.}
    \label{fig:baselines}
\end{figure*}

\subsection{Structure-aware Adaptive Replacement}
\label{sec:adaptive}
During inference, we erase target characters by replacing their corresponding embeddings with the optimized anchor embeddings, ensuring minimal impact on unrelated elements. To further enhance overall scene coherence, we propose a structure-aware adaptive strategy that dynamically introduces the embedding replacement process during image generation.

\noindent \textbf{Embedding Replacement.} When a prompt with $N$ words (including $n$ target-related terms that are pre-listed in a subspace associated with copyrighted characters) is fed into the model's text encoder, it is tokenized and encoded into contextualized embeddings:
\begin{equation}
\label{eq:optimized embedding}
    e = \{e_{SOT}, \> e_{w_1},\> ...,\>e_{w_N},\> e_{EOT},\>e_{Paddings}\}
\end{equation}

For target erasure, we locate all the $n$ target-related embeddings $e_{target}=\{e_{w_{i+1}},...,e_{w_{i+n}}\}$ where $ i\in[0,N-n]$, and replace them with the optimized anchor embedding:
\begin{equation}
    e^{'}_{target}=\{e^*_{anchor},\>...,\>e^*_{anchor}\}
\end{equation}

Furthermore, based on findings that special embeddings (EOT and Paddings) also encode meaningful layout and semantic information~\cite{Brown2020LanguageMA,Zhuang2024MagnetWN}, to enhance anchor concept integration while preserving overall context, we adapt the semantic additivity principle to fuse the special input embeddings $\{e_{EOT},e_{Paddings}\}$ and optimized embeddings $\{e^*_{EOT},e^*_{Paddings}\}$ via element-wise addition~\cite{Hu2024TokenMF}:
\begin{equation}
    e^{'}_{EOT}=\lambda_1 \cdot e_{EOT}\>+\>\lambda_2 \cdot e^*_{EOT}
\end{equation}
\begin{equation}
    e^{'}_{Paddings}=\lambda_1 \cdot e_{Paddings}\>+\>\lambda_2 \cdot e^*_{Paddings}
\end{equation}
where $\lambda_1$ and $\lambda_2$ are determined empirically. 

Final contextualized embeddings for image generation are:
\begin{equation}
\label{eq:final}
    e^{'}=\{e_{SOT},\>e_{w_1},\>…, \> e^*_{anchor},\>…,\>e^{'}_{EOT},\>e^{'}_{Paddings}\}
\end{equation}
This embedding-level replacement avoids modifications to the pretrained text encoder's parameters.

\noindent \textbf{Structure-aware Adaptive Strategy.} To enhance image coherence, we introduce a structure-aware adaptive replacement strategy. During denoising, we apply low-frequency filter $f_L$ at each timestep $t$ to extract structural component $f_L(x_t)$ from the predicted sample. We compute $L_2$ distance between consecutive component:
\begin{equation}
     L_2=||f_L(x_{t})-f_L(x_{t-1})||^2_2,\>\>t\in[1,1000]
\end{equation}

We show a denoising example in Figure~\ref{fig:adaptive_replace}. In this example, $L_2$ increases from initial timestep $t=1000$ to $t\approx720$, reflecting a consistent transformation of the overall image layout. After $t=720$, $L_2$ begins to decrease, indicating structural stabilization and a shift from overall outline toward finer content refinement. This transition point serves as an optimal starting point for embedding replacement, as the stabilized layout provides a reliable spatial reference frame, allowing targeted modifications to specific regions without affecting the established global structure.

Based on the above analysis of structural dynamics during denoising, we propose an structure-aware adaptive replacement strategy. During inference, our method allows the diffusion model to denoise conditioned on the input prompt while simultaneously tracking the variation of low-frequency structural signals.  Embedding replacement is triggered automatically when structural change ceases to increase consistently and reaches a predefined threshold.

\section{Experiments}
\label{sec:experiments}

\subsection{Experimental Setup}
\noindent \textbf{Baselines.} We compare our method with five prompt-based steering baselines, including Safe Latent Diffusion (SLD)~\cite{Schramowski2022SafeLD}, STG~\cite{na2025trainingfree}, SAFREE~\cite{DBLP:conf/iclr/YoonYPYB25}, TraSCE~\cite{Jain2024TraSCETS}, and Negative Prompting (NP). We further evaluate the effectiveness of our optimized anchor embeddings by integrating them into four representative model modification baselines: MACE~\cite{Lu2024MACEMC}, UCE~\cite{Gandikota2023UnifiedCE}, ESD-u~\cite{Gandikota2023ErasingCF}, and AC~\cite{Kumari2023AblatingCI}.

\noindent \textbf{Dataset.} We propose a dataset of 80 animation characters that can be reliably generated by diffusion models. The dataset comprises 36 anthropomorphic characters, 23 animal-form characters, and 21 characters in miscellaneous categories. For evaluation, we generate 100 images for each character using prompts from GPT-4o~\cite{Achiam2023GPT4TR}. 

\noindent \textbf{Target Models.} We utilize the pre-trained Stable Diffusion-v1-4 ~\cite{Rombach2021HighResolutionIS} for all experiments. In addition, we also conduct experiments on Stable Diffusion-v1-5, Stable Diffusion-v2, Stable Diffusion-v2-1, Stable Diffusion-XL-base-1, and Z-Image~\cite{team2025zimage} (DiT-based model) to demonstrate our method's transferability across model architectures. Our method requires no model fine-tuning, so the hyper-parameters of all the pretrained models remain unchanged. 

\noindent \textbf{Evaluation Metrics.} We assess our method from three aspects. For erasure effectiveness of target animation characters, we employ two vision-language models, LLaVA-1.5~\cite{Liu2023ImprovedBW} and BLIP-3~\cite{Lu2024RGBTTV}, to identify the presence of targets. A lower identification accuracy indicates a more complete erasure. For image fidelity preservation, we employ three metrics: SSIM~\cite{1284395} measures the structural similarity; LPIPS~\cite{Zhang2018TheUE} evaluates perceptual difference; Aesthetic Predictor V2 Score ($Aesthetic$)~\cite{LAION-AES} assesses the visual appeal of the erased version. For impact on normal image generation, we generate images from irrelevant prompts in COCO-30K dataset ~\cite{Lin2014MicrosoftCC}, and calculate the Frechet Inception Distance ($FID$)~\cite{Heusel2017GANsTB} and the CLIP Score ~\cite{Radford2021LearningTV}.


\subsection{Quantitative Comparison}
Quantitative comparison results are reported in Table~\ref{tab:baselines experiment}.

\noindent \textbf{Erasure Effectiveness of Target Concept.} Compared to all baselines, images erased using our method achieve the lowest accuracies for successful target identification by both LLaVA-1.5 (6.0\%) and BLIP-3 (4.0\%), with reductions of 3.5\% and 1.7\% compared to the second lowest, respectively. This demonstrates our method's erasure effectiveness, as it effectively deceive multi-modal large models. 

\noindent \textbf{Image Fidelity Preservation.} Our method achieves the highest SSIM of 0.467 and the lowest LPIPS of 0.505 between the original and erased image versions. These results demonstrate our method's superior ability to maintain the integrity and consistency of unrelated concepts and background. Besides, our method achieves the highest Aesthetic score of 5.18, indicating that the erased images possess higher artistic quality. This ensures that the images retain high application value even after target erasure.

\noindent \textbf{Impact on Unrelated Image Quality.} Our method maintains near-identical FID and CLIP Score to the original SD v1.4 (33.7 and 0.326), demonstrating undiminished normal generation capability.

\begin{table}[t!]
\caption{Ablation study on key modules of our method. $\mathbf{\Delta}$ denotes the difference compared to our complete method.}
{\Large
\renewcommand{\arraystretch}{1.25}
\centering
\resizebox{0.96\columnwidth}{!}{
\begin{tabular}{ccccc}
\toprule[1.2pt]
\multirow{2}{*}{\textbf{\Large Method}}  & \multicolumn{2}{c}{\textbf{Erasure Effectiveness}} & \multicolumn{2}{c}{\textbf{Fidelity Preservation}} \\ 
\cmidrule(lr){2-3} \cmidrule(lr){4-5}
                   & LLaVA-1.5↓ ($\mathbf{\Delta}$)         & BLIP-3↓ ($\mathbf{\Delta}$)         & SSIM↑ ($\mathbf{\Delta}$)               & LPIPS↓ ($\mathbf{\Delta}$)              \\ \specialrule{0.9pt}{0.5ex}{0.5ex}
                   SD v1.4 (Base)&66.9\% &64.7\% &1 &0 \\ \hline
w/o Structural Construction& 55.0\% (\textbf{+49.0\%})            & 46.0\% (\textbf{+42.0\%})          & 0.598 (\textbf{+0.131})               & 0.384 (\textbf{-0.121})                \\
w/o Detailed Differentiation& 21.0\% (\textbf{+15.0\%})            & 15.0\% (\textbf{+11.0\%})          & 0.417 (\textbf{-0.050}) & 0.653 (\textbf{+0.148})\\
 w/o Adaptive Replacement & 3.0\% (\textbf{-3.0\%}) & 1.7\% (\textbf{-2.3\%}) & 0.319 (\textbf{-0.148}) &0.668 (\textbf{+0.163})\\
w/o Special Embeddings Addition& 12.7\% (\textbf{+6.7\%}) &                 9.3\% (\textbf{+5.3\%}) & 0.596 (\textbf{+0.129}) & 0.414 (\textbf{-0.091})\\ \rowcolor{blue!10}
Ours                     & 6.0\%            & 4.0\%          & 0.467                & 0.505                \\ 
\bottomrule[1.2pt]
\end{tabular}}}
\label{tab:ablation}
\end{table}

\begin{figure}[t!]   
  \centering  
  \includegraphics[width=0.95\columnwidth]{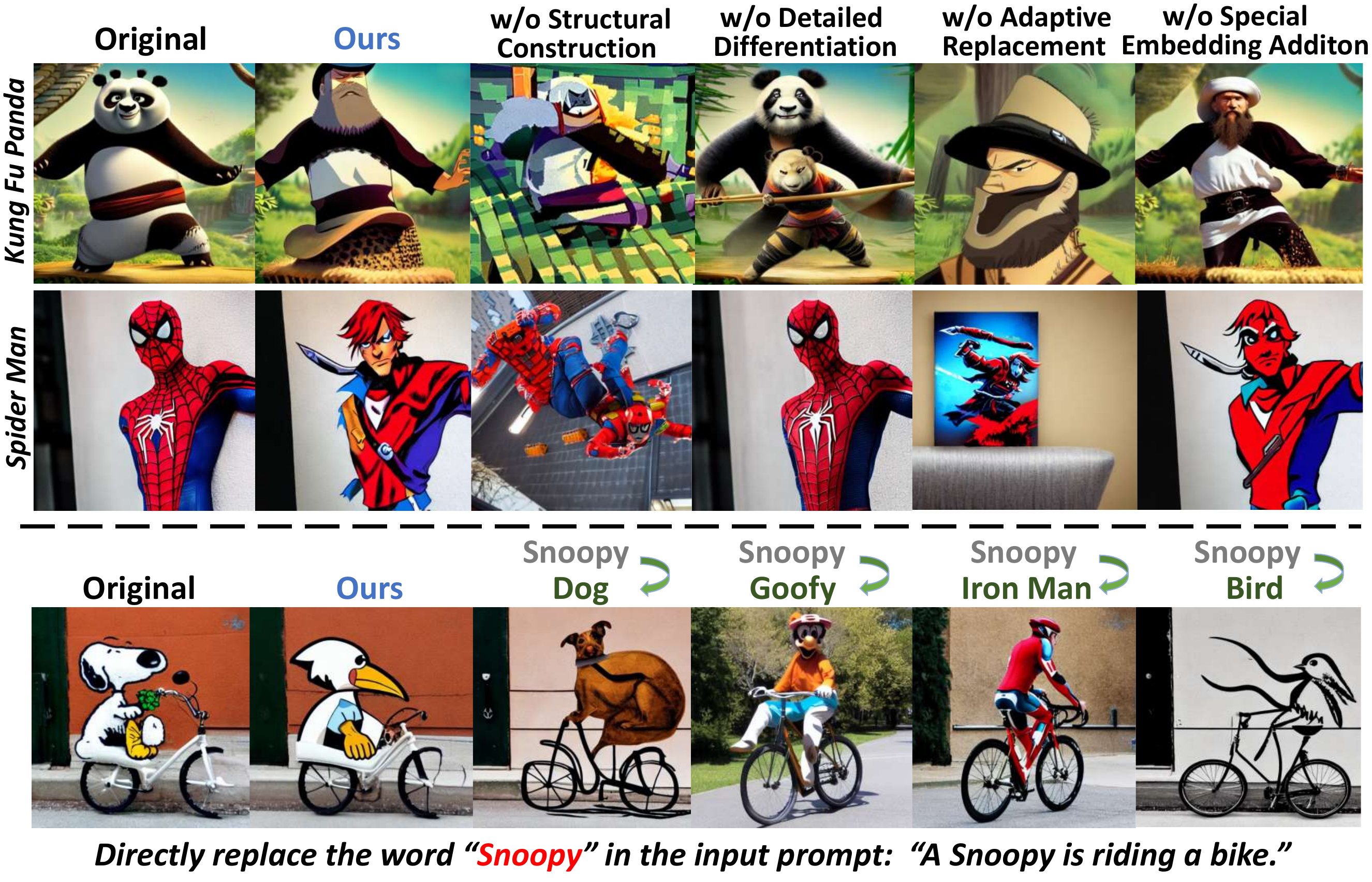} 
  \caption{Visualization of ablation results on key modules.}  
\label{fig:ablation}
\end{figure}

\subsection{Qualitative Comparison}
As shown in Figure~\ref{fig:baselines}, our method achieves seamless erasure of animation characters through optimized anchors, while prompt-based baselines often fail—particularly for characters with complex shapes and attributes (e.g., Donald Duck and Super Mario). Besides, our method achieves background consistency without the blurring or warping artifacts, supporting iterative creative workflows.


\begin{figure}[h!]   
  \centering  
  \includegraphics[width=.9\columnwidth]{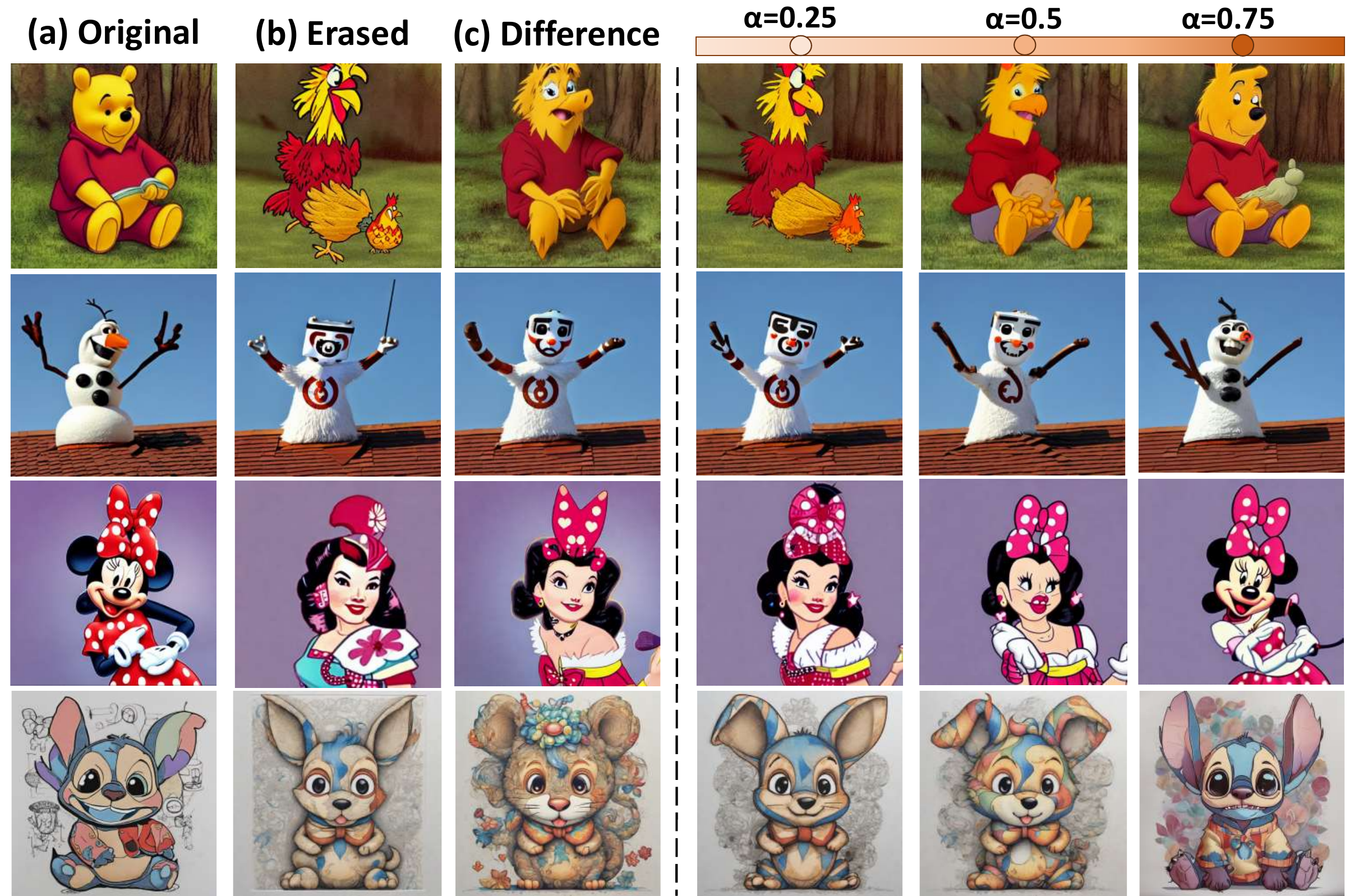} 
  \caption{\textbf{Left:} original images (target), erased version (anchor), and their difference (erased semantic features). \textbf{Right:} results of fine-grained erasure control via $\alpha$-interpolation.}  
\label{fig:finegrained}
\end{figure}

\begin{figure}[h!]   
  \centering  
  \includegraphics[width=.9\columnwidth]{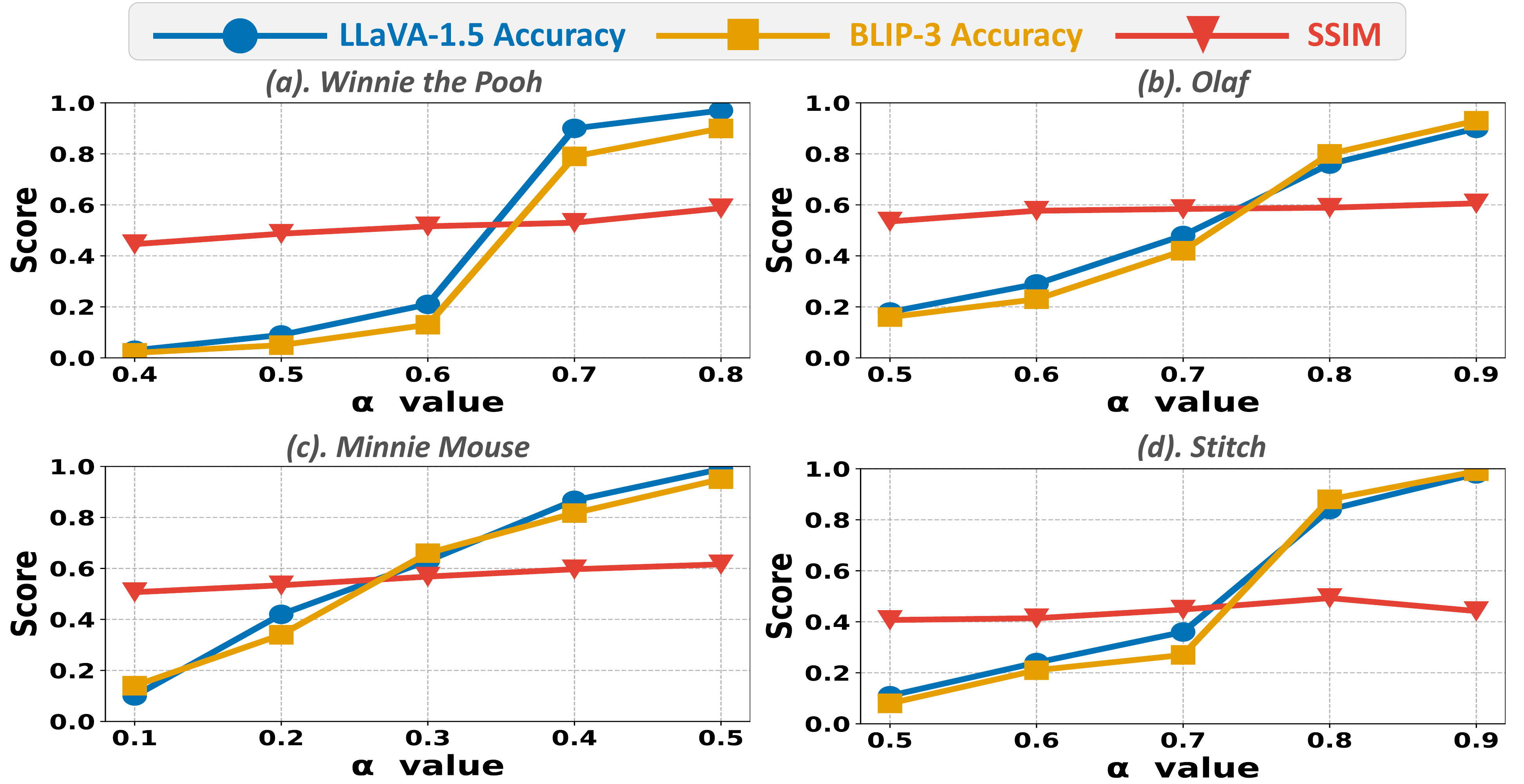} 
  \caption{Performance variation curves under various erasure degrees. Blue and orange lines represent the identification accuracy by LLaVA-1.5 and BLIP-3, while the red line denotes the SSIM value between the erased and original versions.}  
\label{fig:finegrained_curve}
\end{figure}

\subsection{Ablation Study}
We conduct ablation study on our proposed method, and the results are presented in Table~\ref{tab:ablation} and Figure~\ref{fig:ablation}.

\noindent \textbf{Structural and Detailed Modules.} We optimize anchor concepts through joint structural and detailed constraints. To analyze their individual role, we ablate the two modules respectively. Results in Table~\ref{tab:ablation} indicate that both modules are essential: (i) without structural constraints, anchor optimization fails, rendering the replacement ineffective; (ii) without detailed constraints, anchor concepts closely resemble target concepts, degrading erasure performance.

\noindent \textbf{Adaptive Replacement Module.} We propose the adaptive replacement module to better preserve overall structural coherence. To highlight its importance, we present ablation results in Table~\ref{tab:ablation}, where target embeddings are replaced directly from the initial denoising step. While effective target erasure can be achieved, this brings severe structural changes and distortions in the overall image, resulting in significantly reduced image fidelity.

\noindent \textbf{Special Embeddings Addition.} During target replacement, we add the optimized special embeddings (EOT and Paddings) to the original ones. Table~\ref{tab:ablation} reveals that omitting these embeddings reduces erasure effectiveness, indicating that target-related semantics persist in special embeddings and require explicit handling. Thus, adding the optimized special embeddings facilitates better integration of the anchor's semantics while erasing the target's.

\noindent \textbf{Prompt-level Words Modification.} While our method works in the embedding space, a naive alternative is prompt-level word replacement. As shown in the third row of Figure~\ref{fig:ablation}, directly replacing words often causes large layout/style changes. Moreover, structural distortion worsens with increasing semantic distance between target and replacement words. Consequently, naive prompt-level modification lacks the fine-grained control necessary for nuanced animation character removal in practical scenarios. 

\subsection{Fine-grained Control over Erasure Degrees}
Our method effectively achieves fine-grained control over the erasure degree of target characters. Since our anchor concepts are constructed in the continuous embedding space, the erasure degree can be precisely modulated through vector arithmetic. For a target concept embedding $e_{target}$ and its optimized anchor $e_{anchor}$, Figure~\ref{fig:finegrained}(a)(b)(c) illustrate the images generated from $e_{target}$, $e_{anchor}$, and $e_{target}-e_{anchor}$, respectively. Figure~\ref{fig:finegrained}(c) visualizes the semantic features that are erased from the target. 

\begin{figure}[t!]   
  \centering  
  \includegraphics[width=.92\columnwidth]{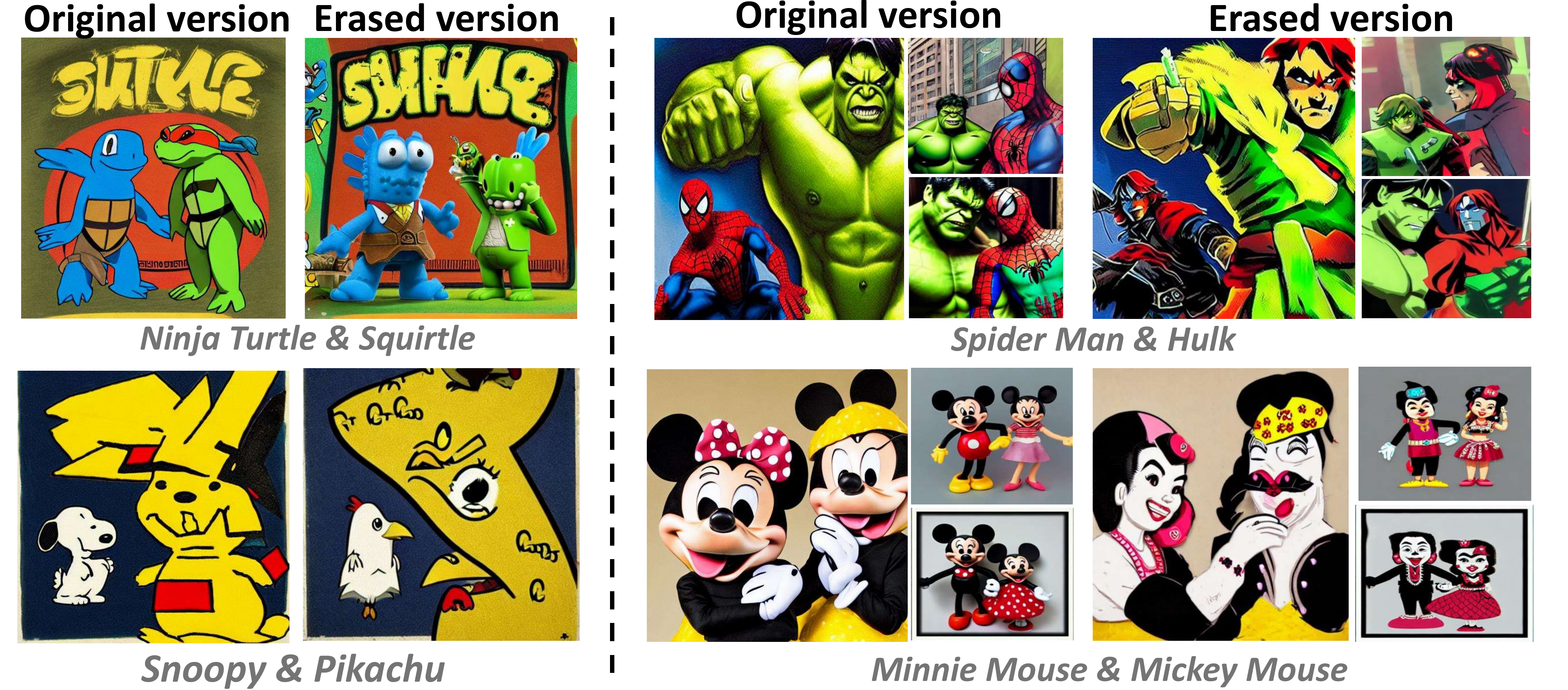} 
  \caption{Simultaneous erasure of multiple characters.}  
\label{fig:Multi_object}
\end{figure}

\begin{table}[t!]
\caption{Transferability of our method across different model versions. Models sharing the same color indicate that they adopt the same text encoder architecture.}
{\Large
\renewcommand{\arraystretch}{1.3}
\centering
\resizebox{0.95\columnwidth}{!}{
\begin{tabular}{ccccccc}
\toprule[1.2pt]
\multicolumn{3}{c}{\textbf{Model Versions}}  & \multicolumn{2}{c}{\textbf{Erasure Effectiveness}} & \multicolumn{2}{c}{\textbf{Fidelity Preservation}} \\ 
\cmidrule(lr){1-3} \cmidrule(lr){4-5} \cmidrule(lr){6-7}
Model     & Backbone            & Dimension             & LLaVA-1.5↓ & BLIP-3↓ & SSIM↑& LPIPS↓\\ \specialrule{0.9pt}{0.5ex}{0.5ex}\rowcolor{blue!10}
SD v1.4 & UNet & 768 & 6.0\% &4.0\% & 0.467 & 0.505 \\ \rowcolor{blue!10}
SD v1.5 & UNet & 768                   & 7.0\%            & 6.0\%         & 0.509                & 0.487                \\ \hline \rowcolor{yellow!10}
SD v2 & UNet  & 1024                  & 8.7\%             &  6.3\%               & 0.598                & 0.437                \\\rowcolor{yellow!10}
SD v2.1 & UNet & 1024                  & 7.7\%              & 5.3\%               & 0.585                & 0.413                \\ \hline \rowcolor{magenta!10}
SDXL & UNet & 768$\>$$\&$$\>$1280                   & 5.6\%               & 4.9\%      & 0.439                & 0.517\\ \hline \rowcolor{green!10}
Z-Image & DiT & 2560                   & 7.9\%               & 7.1\%      & 0.392                & 0.563\\
\bottomrule[1.2pt]
\end{tabular}}}
\label{tab:transferbility}
\end{table}

Continuous fine-grained control is achieved through: 
\begin{equation}
    e^{'}=e_{anchor}+\alpha \cdot (e_{target}-e_{anchor}), \>\> \alpha \in[0,1]
\end{equation}
This formulation interpolates between anchor embedding $e_{anchor}$ (complete erasure) and target embedding $e_{target}$ (no erasure), where $\alpha$ governs the interpolation strength. Increasing $\alpha$ produces images closer to the original, and vice versa. Figure~\ref{fig:finegrained} shows the generated images for $\alpha=0.25,0.5,0.75$, respectively.

We plot the variation of identification accuracy (LLaVA-1.5 and BLIP-3) and SSIM with $\alpha$ for four animation characters in Figure~\ref{fig:finegrained_curve}. The stable SSIM values confirm that image structure is largely unaffected by $\alpha$. In contrast, identification accuracy differs markedly: Winnie the Pooh and Stitch exhibit a sharp increase within a small $\alpha$ interval, while Olaf and Minnie Mouse improve more gradually. Minnie Mouse attains high accuracy at smaller $\alpha$, likely due to its highly unique and recognizable features, whereas the others require larger $\alpha$ (especially Stitch). This enables flexible and user-tailored control over the target erasure degrees for different characters.

\subsection{Simultaneous Erasure of Multi-targets}
Benefiting from the precise localization of target concepts and the feasibility of multi-embeddings replacement, our method can also achieve simultaneous erasure of multiple targets. First, we optimize an anchor embedding for each animation character. Subsequently, we locate all the target-related embeddings and simultaneously replace them with their corresponding anchor embeddings, ensuring visual coherence. Results are presented in Figure~\ref{fig:Multi_object}.

\subsection{Transferability across Different Models}
Our method is transferable across different diffusion models. The optimized anchor embedding enables seamless deployment on any diffusion model sharing the same text encoder architecture (with the same embedding dimension $1\times D$), thus avoiding redundant optimization. For instance, anchor embeddings optimized in SD v1.4 can be used in SD v1.5 ($D=768$), and those for SD v2 and SD v2.1 can be shared ($D=1024$). Moreover, our method is also applicable to models with dual text encoders, such as SDXL ($D=768 \>\&\> 1280$), by jointly optimizing the anchor embeddings in both text encoders. Cross-model results (Table~\ref{tab:transferbility}) highlights the versatility of our method.

\noindent \textbf{Transferability to DiT-based Models.} As shown in Table~\ref{tab:transferbility}, we effectively extend our method to Z-Image~\cite{team2025zimage}, a recently released DiT-based model with flow-matching~\cite{Lipman2022FlowMF,Liu2022FlowSA} sampling strategy. 

\begin{table}[]
\caption{Quantitative results of integrating our optimized anchors into existing model modification baselines, compared with using general anchors (i.e. null text or ``toy'').}
{\Large
\renewcommand{\arraystretch}{1.25}
\centering
\resizebox{0.95\columnwidth}{!}{
\begin{tabular}{ccccccc}
\toprule[1.2pt]
\multicolumn{3}{c}{\textbf{Method}}       & \multicolumn{1}{c}{\textbf{Erasure Effectiveness}} & \multicolumn{3}{c}{\textbf{Fidelity Preservation}} \\
\cmidrule(lr){1-3} \cmidrule(lr){4-4} \cmidrule(lr){5-7}
Baseline      &Fine-tuning         & Anchor   & LLaVA-1.5↓            & SSIM↑          & LPIPS↓         & Art↑          \\ \specialrule{0.9pt}{0.5ex}{0.5ex}
\multirow{2}{*}{UCE} & \multirow{2}{*}{\textcolor{red!70!black}{\ding{55}}} & General & 9.6\%& 0.248& 0.698& 4.74\\
          &             & \cellcolor{blue!10}Ours    & \cellcolor{blue!10} \textbf{7.2\%}& \cellcolor{blue!10} \textbf{0.382}& \cellcolor{blue!10} \textbf{0.669}& \cellcolor{blue!10} \textbf{4.94}\\ \hline
 \multirow{2}{*}{ESD-u} & \multirow{2}{*}{\textcolor{green!70!black}{\ding{51}}} & General& 36.0\%& 0.315& 0.688&4.83\\ &
 & \cellcolor{blue!10} Ours& \cellcolor{blue!10} \textbf{13.6\%}& \cellcolor{blue!10} \textbf{0.358}& \cellcolor{blue!10} \textbf{0.640}& \cellcolor{blue!10} \textbf{4.99}\\ \hline
\multirow{2}{*}{AC} & \multirow{2}{*}{\textcolor{green!70!black}{\textcolor{green!70!black}{\ding{51}}}}   & General & 6.0\%                 & 0.268          & 0.698          & 4.74          \\
           &            & \cellcolor{blue!10} Ours    & \cellcolor{blue!10} \textbf{3.2\%}        & \cellcolor{blue!10} \textbf{0.315} & \cellcolor{blue!10} \textbf{0.672} & \cellcolor{blue!10} \textbf{5.01} \\ \hline
\multirow{2}{*}{MACE} & \multirow{2}{*}{\textcolor{green!70!black}{\ding{51}}} & General & 9.0\%                 & 0.282          & 0.720          & 4.76          \\
            &           & \cellcolor{blue!10} Ours    & \cellcolor{blue!10} 17.5\%                & \cellcolor{blue!10} \textbf{0.302} & \cellcolor{blue!10} \textbf{0.677} & \cellcolor{blue!10} \textbf{4.76 }         \\ 
\bottomrule[1.2pt]
\end{tabular}}}
\label{tab:tuning}
\end{table}

\subsection{Anchor Integration into Model Modification}
For each character, our method optimizes its anchor embedding in the text encoder, denoted as ``Anchor*''. While this anchor is primarily used for inference-time replacement in our main pipeline, it can also be viewed as a plug-and-play component for existing model modification baselines. Specifically, we integrate our optimized anchor by simply substituting the baselines' original modification targets with ``Anchor*''. As shown in Table~\ref{tab:tuning}, compared to using semantically distant general anchors (e.g., null text or ``toy''), our semantically proximate anchors generally improve erasure effectiveness and better preserve image fidelity across most baselines, without altering any other method operations.

\section{Conclusion}
\label{sec:conclusion}
In this paper, we address animation copyright infringement by proposing a controllable method to erase animation characters during diffusion-based image generation. We first optimize an anchor in the continuous embedding space under structural and detail constraints, then replace target-related embeddings with the learned anchor via a structure-aware adaptive strategy. Experiments demonstrate our method's state-of-the-art erasure accuracy, image fidelity preservation, and support for controllable erasure degree, multi-target removal, and model transferability. We hope our contributions not only help regulators prevent infringement but also maximally preserve users’ creative intent, facilitating deployment of trustworthy and user-centric AI systems.

\section{Acknowledgments}
This work was supported by the National Key Research and Development Program of China (No.2024YFC3307402).

\bibliographystyle{ACM-Reference-Format}
\balance
\bibliography{acmart}

\end{document}